\pdfoutput=1

\documentclass[11pt]{article}

\usepackage[]{acl}

\usepackage{times}
\usepackage{latexsym}
\usepackage{amsmath}
\usepackage{amssymb}
\usepackage{bm}
\usepackage{multirow}
\usepackage{multicol}
\usepackage{booktabs}
\usepackage{xspace}
\usepackage{amsfonts}
\usepackage{xcolor}
\usepackage{bbm}
\usepackage{tabularx}
\usepackage{graphicx}
\usepackage[textsize=scriptsize]{todonotes}

\definecolor{green1}{HTML}{579D42}
\definecolor{blue1}{HTML}{005F86}

\newcommand{\liyan}[1]{\textcolor{black}{\textbf{}#1}}

\usepackage[T1]{fontenc}

\usepackage[utf8]{inputenc}

\usepackage{microtype}

\title{Less Likely Brainstorming: Using Language Models \\to Generate Alternative Hypotheses}

\author{Liyan Tang$^\diamondsuit$ \quad Yifan Peng$^\spadesuit$ \quad Yanshan Wang$^\clubsuit$ \quad Ying Ding$^\diamondsuit$ \\ \bf Greg Durrett$^\diamondsuit$ \quad Justin F. Rousseau$^\diamondsuit$ \\
        $^\diamondsuit$The University of Texas at Austin \\ $^\spadesuit$Weill Cornell Medicine \quad $^\clubsuit$University of Pittsburgh \\
\texttt{lytang@utexas.edu}
}

\begin{document}
\maketitle
\begin{abstract}
A human decision-maker benefits the most from an AI assistant that corrects for their biases.
For problems such as generating interpretation of a radiology report given findings, a system predicting only highly likely outcomes may be less useful, where such outcomes are already obvious to the user. To alleviate biases in human decision-making, it is worth considering a broad differential diagnosis, going beyond the most likely options. We introduce a new task, ``less likely brainstorming,'' that asks a model to generate outputs that humans think are relevant but less likely to happen. We explore the task in two settings: a brain MRI interpretation generation setting and an everyday commonsense reasoning setting. We found that a baseline approach of training with less likely hypotheses as targets generates outputs that humans evaluate as either likely or irrelevant nearly half of the time; standard MLE training is not effective. To tackle this problem, we propose a controlled text generation method that uses a novel contrastive learning strategy to encourage models to differentiate between generating likely and less likely outputs according to humans. We compare our method with several state-of-the-art controlled text generation models via automatic and human evaluations and show that our models' capability of generating less likely outputs is improved.\footnote{Code is available at \url{https://github.com/Liyan06/Brainstorm}.} 
\end{abstract}

\section{Introduction}

Cognitive errors occur when an abnormality is identified, but its importance is incorrectly understood, resulting in an incorrect final diagnosis \cite{Onder2021, Bruno2015}. For example, radiologists may look for confirmatory evidence to support a diagnostic hypothesis and ignore or discount evidence that refutes the hypothesis (confirmation bias; \citet{Busby2018, Onder2021}). One way to reduce the likelihood of such cognitive errors is to provide cognitive ``help'' by having a devil’s advocate \cite{Seah2021, Waite2017}. For this purpose, we propose a new text generation task called ``\textbf{less likely brainstorming}'' to produce less likely but relevant consultations to bring fresh eyes to examine a case---a powerful way to correct diagnostic errors.


\begin{figure}
    \centering
    \includegraphics[width=\linewidth, trim=0mm 0mm 00mm 0mm,clip]{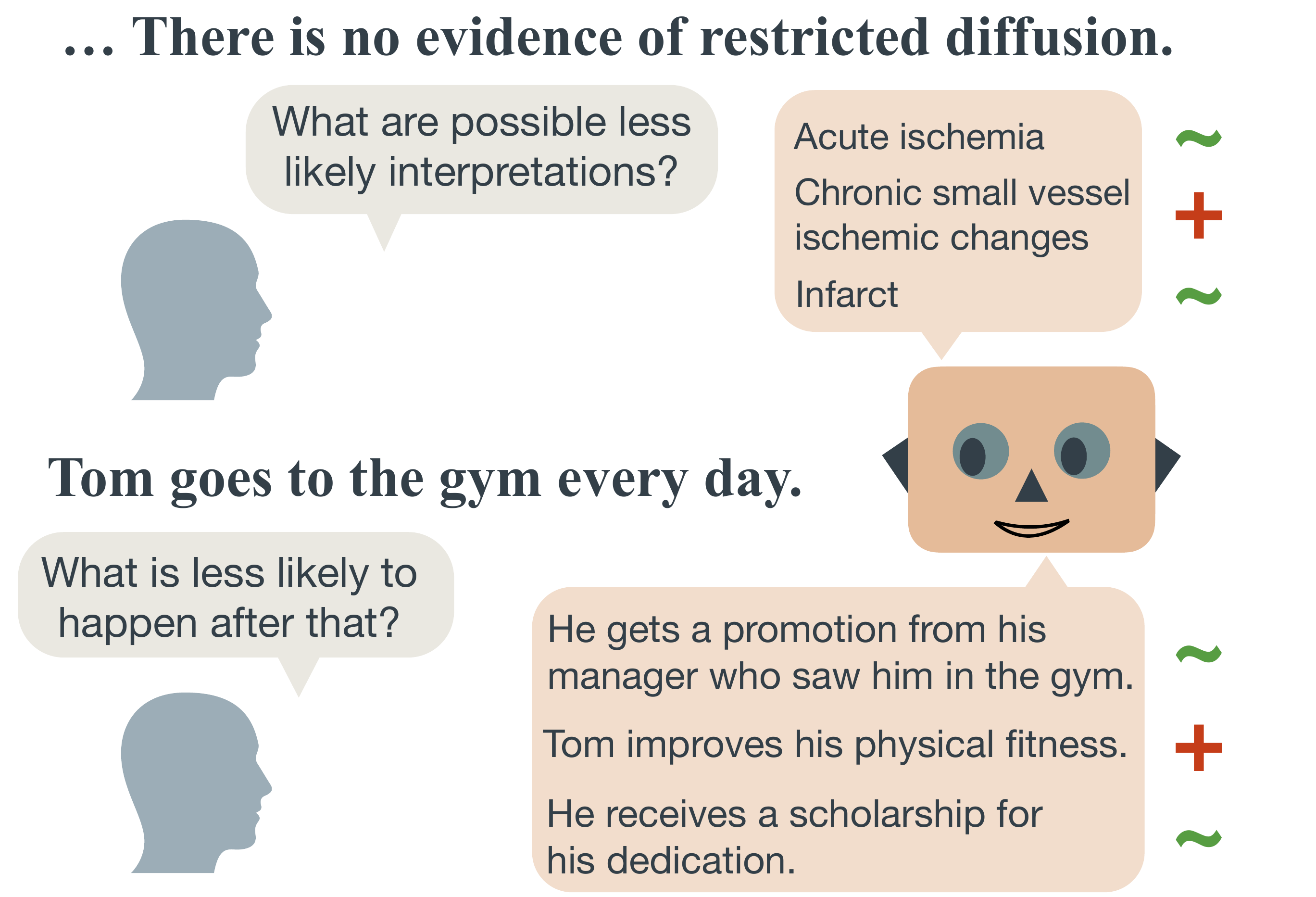}
    \caption{Examples from \textsc{MRIInterpret} and E-CARE datasets. The task is to generate interpretations or hypotheses that humans would consider to be ``less likely'' to happen but still relevant to the context. ``$+$'' and ``$\sim$'' represent likely and less likely outputs, respectively.}
    \label{fig:illustration}
\end{figure}


Here, we consider less likely hypotheses in two scenarios. First, they can be hypotheses that humans think are likely but not among the most likely to happen. These hypotheses are critical to providing second opinion of a prior clinical study but are often difficult to generate by traditional decoding techniques. Second, they can be hypotheses that \emph{are} indeed impossible according to humans, but are close to being true if certain counterfactual assumptions about the input hold. These hypotheses are also helpful as they are often ignored by clinicians. There is a tendency for clinicians to look for a confirmatory diagnostic hypothesis but ignore a refutable one.
Note that a less likely hypothesis reflects the likelihood of a potential diagnosis \emph{from the human perspective}, not from the probability of model output.

We propose \textsc{Brainstorm}, a novel contrastive learning strategy to generate ``less likely'' hypotheses. We treat this problem as a text generation task as text generation models are the most flexible for providing predictions and explanations for complex tasks; they can generalize to new examples and produce complex, structured diagnoses in many formats. Generation of the ``less likely hypotheses'' is conditioned on an indicator variable set to trigger the model to prefer outputs are less likely according to humans.
For this purpose, we propose two additional loss objectives to effectively learn the relationship between input context, the indicator, and outputs. 
Without our training strategy, using naive controlled generation training, we find that conditioning on the indicator often leads to generating ``highly likely'' or irrelevant outputs.


We explore this task in two settings: everyday commonsense reasoning and brain magnetic resonance imaging (MRI) interpretation generation \liyan{(more details in Section~\ref{sec:experiment})}. In the everyday commonsense reasoning setting, we adapt $\textsc{Art}$ \cite{Bhagavatula2020Abductive} and E-CARE \cite{du-etal-2022-e}, which both contain ``less plausible'' or ``implausible'' hypotheses that fit our definition of less likely. \liyan{An illustrative example asking for less likely hypotheses can be found in Figure~\ref{fig:illustration}.} We show that our approach can generate more ``less likely'' hypotheses than baselines, including models directly fine-tuned on this set, past controllable generation approaches  \cite{Lu2022QuarkCT}, or models with alternate decoding \cite{Li2022, liu-etal-2021-dexperts}. In the brain MRI interpretation setting, we experiment with predicting diagnoses from brain MRI reports \liyan{(see Figure~\ref{fig:illustration})}. Assessment by a neurologist reveals that our model successfully shifts the distribution of generated diagnoses further toward the tail while still generating relevant diagnoses.

\section{Related Work} \label{sec:related_work}

\paragraph{Uncertainty in Radiology Interpretation}

Uncertainty plays a significant role in the process of clinical decision making \cite{Croskerry2013}. When facing uncertainty, physicians may resort to various erroneous strategies, such as denying the presence of uncertainty resulting in various interpretation biases. These biases could lead to unexpected consequences \cite{Kim2018, Eddy1984}, including missed diagnoses, misdiagnoses, unnecessary diagnostic examinations and even life-threatening situations \cite{Farnan2008}. Recent work \cite{Seah2021, Waite2017} have provided deep-learning based methods and suggestions in reducing errors from interpretation bias on medical imaging. To the best of our knowledge, we are the first to explore reducing bias from interpreting radiology reports via our less likely text generation framework.

\paragraph{Controllable text generation and decoding methods} Controllable text generation is the task of generating text that adheres certain attributes, such as language detoxification \cite{DisCup-2022, liu-etal-2021-dexperts, Dathathri2020Plug}, formality modification \cite{mireshghallah-etal-2022-mix, yang-klein-2021-fudge} and open-ended story generation \cite{mori-etal-2022-plug, lin-riedl-2021-plug, fan-etal-2018-hierarchical}. The task of controllable text generation encompasses both training-time and decoding-time methods. Training-time approaches include CTRL \cite{Keskar2019CTRLAC}, which learns to utilize control codes to govern attributes in order to generate the desired text, and \textsc{Quark} \cite{Lu2022QuarkCT}, which leverages a strong attribute classifier as a reward function to unlearn unwanted attributes. These methods typically rely on training data that contains both the desired and undesired attributes to be effective in the supervised setting. Our method falls into this category. 

On the other hand, decoding-time methods utilize off-the-shelf pre-trained LMs (PLMs) and aim to re-rank the probability of generated text based on specific constraints. PPLM \cite{Dathathri2020Plug} and FUDGE \cite{yang-klein-2021-fudge} are typical methods in this category that train an attribute classifier to guide PLMs to generating desired text. \textsc{DExperts} \cite{liu-etal-2021-dexperts} and Contrastive Decoding \cite{Li2022} are more recent methods that re-weight generation probabilities by contrasting the output distributions between different LMs. We select those two as strong baselines for comparison against our proposed model. 



\paragraph{Contrastive Learning in NLP}

Contrastive learning (CL) has been applied to a wide range of representation learning tasks in NLP, such as learning task-agnostic sentence representation \cite{gao-etal-2021-simcse} and improving natural language understanding \cite{pmlr-v158-jaiswal21a, qu-etal-2021-coda}. It has recently been applied to text generation tasks as well \cite{An-etal-2022-cont, cao-wang-2021-cliff, Lee-etal-2021-contrast} where additional hard positive or negative examples are created through techniques such as back-translation or perturbation.





\section{Problem Setting}

\liyan{The problem we tackle in this work can be viewed as a controllable text generation task. Let $x$ be a premise or a brain MRI report findings, we want a model to generate a likely/less likely hypothesis or interpretation $y$ given an indicator $i$ by drawing from the distribution $P({y} \mid {x}, i)$. The indicator $i$ can take two values: $+$ to indicate generating likely outputs and $\sim$ to generate less likely outputs.}

\liyan{For example, given a premise $x$ =\emph{``Tom goes to the gym every day.''} in Figure~\ref{fig:illustration} from the \textsc{E-CARE} dataset (more details in Section~\ref{sec:experiment}), we want a model to generate a hypothesis $y^\sim$ that is less likely to happen ($i$ = $\sim$) after $x$, such as \emph{``He gets a promotion from his manager who saw him in the gym.''}. Although this hypothesis fits into the same scenario as the premise as it directly connects to the premise involving Tom's daily gym attendance, it is less likely to happen since the causal relationship between going to the gym and receiving a promotion is not common.} The understanding of what is ``less likely'' can be based on the concept of bounded rationality \cite{Simon1955}, where likely hypotheses are those that are likely given known premises, but less likely hypotheses may stem from additional unknown premises. 

It is important to note that when we refer to an output as ``less likely/likely'', we mean that it is less likely/likely based on human understanding of ${x}$. All models we experiment with in this work generate outputs that have high probability according to the model, regardless of whether they are likely or less likely to happen according to humans.




\section{Methodology}

In this section, we present our method as well as baseline models we compare against. Requirements for these models can be found in Table~\ref{tab:model_summary}. We use BART \cite{lewis-etal-2020-bart} as the backbone LM for all experimental settings.

\subsection{\textsc{Brainstorm}}

Our encoder-decoder system takes the concatenation of a pair ($x$, $i$) as input and returns one or multiple generated output sequences $y$.
At decoding time $t$, our model iteratively decodes the next token conditioned on the left-hand context, i.e., $y_{<t}$:
\begin{equation}
    P_{\mathrm{LM}}(y) = \prod_t^T P_{\mathrm{LM}}(y_t \mid x, i, y_{<t})
\end{equation}
where $P_{\mathrm{LM}}(y_t \mid x, i, y_{< t})$ is the next token distribution given the context. 
The task inputs are described in Section~\ref{sec:experiment}.

\begin{figure*}
    \centering
    \includegraphics[width=0.6\linewidth, trim=0mm 0mm 00mm 0mm,clip]{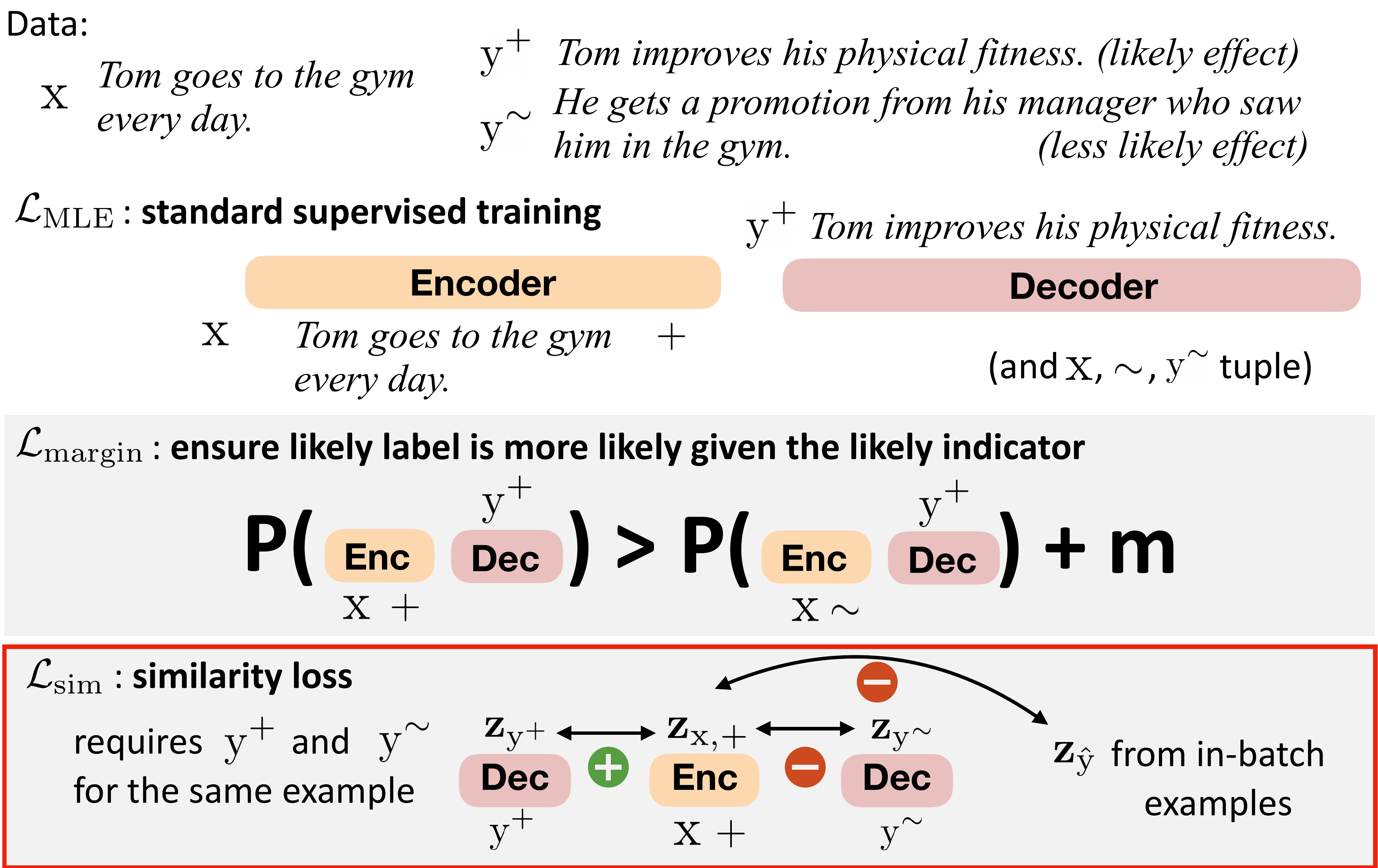}
    \caption{An overview of \textsc{Brainstorm} using an example from \textsc{E-CARE}, which consists of three objectives. $\mathbf{z}_{{x,i}}$ is the encoder representation of the input $x$ conditioned on an indicator $i$. $\mathbf{z}_{{y^+}}, \mathbf{z}_{{y^\sim}}$ and $\mathbf{z}_{\hat{y}}$ are the decoder representations of positive, hard negative, and other negative target sequences within the same batch, respectively. The $\mathcal{L}_{\mathrm{sim}}$ objective is highlighted in red where it requires both likely and less likely data.}
    \label{fig:model_arch}
\end{figure*}

Besides the standard maximum likelihood training with human reference, we incorporate two additional loss objectives to guide models to associate the context, indicators, and target sequences. The training approach is illustrated in Figure~\ref{fig:model_arch}. 

\paragraph{Margin Loss} First, given the indicator $i$, we want the model to assign a higher estimated probability to human reference $y$ than its opposite indicator $\neg i$. Therefore, we apply a margin-based loss:
\begin{equation} \label{equ:margin_loss}
    \mathcal{L_\mathrm{margin}} =  \max (0, P(y \mid x, \neg i) - P(y \mid x, i) + m)
\end{equation}
where $m$ is the margin value. This loss objective tells models that if the indicator is modified, then the target sequence should have lower probability. 
Margin loss does not require both \liyan{likely and less likely outputs} $y^+$ and $y^\sim$.

\paragraph{Similarity Loss}  We propose two versions of a contrastive similarity loss based on the availability of examples that can be used in CL. When both positive and negative examples are available in the same batch, we define the similarity loss as
\begin{equation} \label{loss:sim_loss}
    \mathcal{L_\mathrm{sim}} = - \log \frac{\mathrm{exp}(\mathrm{sim}(\mathbf{z}_{x, i}, \mathbf{z}_y) / \tau)}{\sum_{\hat{y} \in \mathrm{batch}} \mathrm{exp}(\mathrm{sim}(\mathbf{z}_{x, i}, \mathbf{z}_{\hat{y}}) / \tau)}
\end{equation} 
Here, $\mathbf{z}_{x,i}$, $\mathbf{z}_y$, and $\mathbf{z}_{\hat{y}}$ represent the hidden representations of input ($x$, $i$), human reference $y$, and an output $\hat{y}$ in the same batch. $\mathcal{L_\mathrm{sim}}$ encourages the model to maximize the agreement between $\mathbf{z}_{x,i}$ and its corresponding output $\mathbf{z}_y$.  This loss objective encourages a model to learn the relation between certain indicators and the target sequence by contrasting the target sequence with all negative outputs in the batch.

This objective term resembles that in CoNT \cite{An-etal-2022-cont} which takes self-generated outputs as negative samples; here, we conditioned the input on special indicators. Note that at the training time, the indicator $i$ could be either $+$ or $\sim$. When the indicator $i = +$, the hard negative is the human reference of $y^\sim$, and vice versa. We set the weight of the term in Equation (\ref{loss:sim_loss}) associated with the hard negative to $10$ throughout the experiment to increase its importance relative to in-batch negatives.

When positive and negative examples are not available at the same time (denoted by a lack of a ``pair'' check in Table~\ref{tab:model_summary}), we propose an alternative similarity loss objective $\mathcal{L'_\mathrm{sim}}$ that minimizes the similarity of encoder representation $\mathbf{z}_{x, i}$ and $\mathbf{z}_{x, \neg i}$, without comparing to outputs in the batch:
\begin{equation} 
    \mathcal{L'_\mathrm{sim}} = {\mathrm{sim}(\mathbf{z}_{x, i}, \mathbf{z}_{x, \neg i})}.
\end{equation} 
We use cosine similarity for both versions. 



\paragraph{Final Loss} The overall training objective of \textsc{Brainstorm} is the combination of the standard maximum likelihood estimation (MLE) $\mathcal{L_{\mathrm{MLE}}}$, margin loss, and similarity loss:
\begin{equation}
    \mathcal{L_{\mathrm{final}}} = \mathcal{L_{\mathrm{CE}}} + w_s\mathcal{L_\mathrm{sim}} + w_m\mathcal{L_\mathrm{margin}}
\end{equation}
where $w_s$ and $w_m$ are hyperparameters. \textsc{Brainstorm}$'$ replaces $\mathcal{L_\mathrm{sim}}$ by $\mathcal{L'_\mathrm{sim}}$.

\subsection{Baselines}

\subsubsection{Training-Time Baselines}


\paragraph{\textsc{Mle} and \textsc{Mle-LL}}
\textsc{Mle} is trained on all data. It is a conditional model $p(y \mid x, i)$ that learns to generate both $y^+$ and $y^\sim$ depending on $i$. \textsc{Mle-LL} learns to generate less likely outputs $y^\sim$ by only training on $(x, y^\sim)$. Both models are trained with standard MLE.

\paragraph{\textsc{Quark}} \cite{Lu2022QuarkCT} is a state-of-the-art controllable text generation method that outperforms methods such as unlikelihood training \cite{Welleck2020Neural}. \textsc{Quark} trains an LM to generate text with fewer undesirable properties by maximizing rewards assigned by a reward function. In this study, we use the DeBERTa model \cite{He2020DeBERTaDB} as the reward function to help generate more $y^\sim$ (more details in Section~\ref{sec:result}).

\subsubsection{Decoding-Time Baselines} \label{sec:dec_baseline}


\paragraph{Modified \textsc{DExperts}} 
\textsc{DExperts} \cite{liu-etal-2021-dexperts} combines a base LM $M$ along with two language models called ``expert'' ($M_\mathrm{exp}$) and ``anti-expert'' ($M_{\mathrm{anti}}$) that model text with desired and undesired properties, respectively. The next token distribution is determined by
   $P_{\mathrm{DExperts}}(y_t) = \sigma(z'_t + \alpha(z_t^{\mathrm{exp}} - z_t^{\mathrm{anti}}))$
where $z$ is the logits for the next token $y_t$ and $z'_t$ is the truncated logits from $M$ under any truncation sampling methods such as top-$k$ sampling. For simplicity, we omit the preceding context in the notation. The hyperparameter $\alpha$ controls how far the final token distribution deviates from model $M$. 

In our setting, we modify this definition to be
\begin{equation}
    P_{\mathrm{DExperts'}}(y_t) = \sigma(z_t^\sim + \alpha(z_t^{\mathrm{neu}} - z_t^{+}))
\end{equation} 
Here, $z_t^{+}$ is from the model that learns to generate $\hat{y}^+$ by only training on $(x, y^+)$ pairs. $z_t^{\mathrm{neu}}$ is from the model that learns to generate both $y^+$ and $y^\sim$ conditioned on the indicator. Unlike \textsc{Mle}, this model does not condition on indicators to generate hypotheses. Instead, it leverages text with both desired (generating $y^\sim$) and undesired properties (generating $y^+$). It is shown to effectively maintain the fluency of the generated text \cite{liu-etal-2021-dexperts}. $z_t^\sim$ is from a base LM that generates $y^\sim$ only. It can be \textsc{Mle-LL} or \textsc{Brainstorm}.

\paragraph{Modified Contrastive Decoding}
Contrastive Decoding (\textsc{CD}) combines a larger $M_\mathrm{exp}$ and a smaller ``amateur'' model ($M_\mathrm{ama}$) and searches for text under a constrained search space \citep{Li2022}. The resulting outputs are intended to amplify the strengths of $M_\mathrm{exp}$ and remove undesired properties that appear in $M_\mathrm{ama}$. A scaling factor $\tau_\mathrm{CD}$ controls the penalties of the amateur model in CD.

In our setting, two models have the same size. $M_\mathrm{ama}$ learns to generate $y^+$; $M_\mathrm{exp}$ can be \textsc{Mle-LL} or \textsc{Brainstorm}. Intuitively, the ability to generate $y^\sim$ is preserved, while the tendency to generate $y^+$ is factored out.

\begin{table}
\small
\centering
\begin{tabularx}{\columnwidth}{Xccccc}
\toprule
\multirow{2}{*}{\textbf{Methods}} & \multicolumn{3}{c}{\textbf{Data}} & \multicolumn{1}{c}{\multirow{2}{*}{\begin{tabular}[c]{@{}c@{}}\textbf{Need}\\ \textbf{Clf.}\end{tabular}}} \\
\cmidrule{2-4}
& $+$ & $\sim$ & \textbf{pair} \\
\midrule
Training-time Method\\
\hspace{1em}\textsc{Mle-LL} &  & \checkmark &  &  \\
\hspace{1em}\textsc{Mle} &  &  & \checkmark &  \\
\hspace{1em}\textsc{Quark} &  \checkmark&  \checkmark &  \checkmark & \checkmark\\
\hspace{1em}\textsc{Brainstorm} &  &  & \checkmark &  \\
\hspace{1em}\textsc{Brainstorm}$'$ & \checkmark & \checkmark &  &  \\
\midrule
Decoding-time Method\\
\hspace{1em}\textsc{DExperts} &  &  &  \checkmark & \\
\hspace{1em}\textsc{CD} &  &  & \checkmark &\\
 \bottomrule
\end{tabularx}
\caption{Requirements for various methods. $+$/$\sim$/pair means a method requires $y^+$/$y^\sim$/both for $x$. \textsc{Quark} can take any type of data as inputs but requires a trained classifier. We use \textsc{Brainstorm}$'$ as an alternative of \textsc{Brainstorm} if $y^+$ and $y^\sim$ are not both available for $x$. \textsc{DExperts} and \textsc{CD} require that both $y^+$ and $y^\sim$ \emph{could be} available for $x$ (which is not the case for \textsc{MRIInterpret}, Section~\ref{sec:mri_human_eval}).}
\label{tab:model_summary}
\end{table}



\paragraph{Hyperparameters} We experiment with a wide range of values for $\alpha$ in \textsc{DExperts} and $\tau_{\mathrm{CD}}$ in \textsc{CD} and show how the fraction changes across these values in Figure~\ref{fig:anlg_compare}. We keep the recommended value for the remaining hyperparameters. Unless specified otherwise, we generate outputs using diverse beam search \cite{vijayakumar2016diverse}.

\section{Experimental Settings} \label{sec:experiment}


We investigate our methods in both brain MRI settings and everyday commonsense reasoning settings (Table~\ref{tab:stats}).


\subsection{Everyday Commonsense Reasoning} \label{sec:commonsense}

Two datasets from the commonsense reasoning domain were adapted. See examples in Figure~\ref{fig:motivation_examples} from Appendix.

\textbf{\textsc{Art}} (\emph{A}bductive \emph{R}easoning in narrative \emph{T}ext; \citet{Bhagavatula2020Abductive}) is a large-scale benchmark dataset that tests models' language-based abductive reasoning skills over narrative contexts. Each instance in the dataset consists of two observations $O_1$ and $O_2$ ($O_1$ happened before $O_2$), as well as a likely and a less likely hypothesis event (happening in between $O_1$ and $O_2$) collected from crowd workers. 
Each ``likely'' hypothesis is causally related to two observations and each ``less likely'' hypothesis is created by editing each ``likely'' hypothesis. The original task is to generate a likely hypothesis given the observation pair ($O_1$, $O_2$).

\textbf{\textsc{E-CARE}} (Explainable CAusal REasoning;  \citet{du-etal-2022-e}) tests models' causal reasoning skills. Each instance in the dataset consists of a premise, a ``likely'' and a ``less likely'' hypothesis, and a conceptual explanation of the causality. The likely hypothesis can form a valid causal fact with the premise. Two tasks are introduced: (1) causal reasoning: choosing the ``likely'' hypothesis given a premise and (2) explanation generation: generating an explanation for the causal fact. 

\paragraph{Adapted Setting} In our adapted setting, we want a model $F$ to generate $y^\sim$ given either an observation pair (\textsc{Art}) or a premise (\textsc{E-CARE}) $x$. Formally, let $E$ be a binary evaluator $E(x, y) \in \{1, 0\}$ that classifies an output $y$ into either $y^+$ or $y^\sim$ based on $x$. We want a model $F$ that generates $\hat{y} = F(x, i=\sim)$, where $E(x, \hat{y}) = 0$.

\paragraph{Evaluation}

For \textsc{Art}, we use the default training, validation and test sets to evaluate our models. For \textsc{E-CARE}, we randomly construct training and validation sets from the original training set and use the default validation set as the test set since the original test set is not available. All hyperparameters are determined on the validation set. 

For each instance $x$ in the test set, we ask a model $F$ to generate $\hat{y} = F(x, i=\sim)$, then measure the fraction of less likely hypotheses according to an evaluator $E$. 

\liyan{To reduce ambiguity and encourage more consistent human evaluations, we formally define all relevancy categories from rounds of pilot studies. More detailed definitions and annotation instructions can be found in Appendix~\ref{sec:definitions} and~\ref{sec:quality_check}.} We measure both the (1) \emph{relevancy} and (2) \emph{fluency} of generated hypothesis in human evaluation.

\subsection{\textsc{MRIInterpret}}

We present a new dataset \textsc{MRIInterpret} based on the findings and impression sections of a set of de-identified radiology reports we collected from brain MRIs. Each instance consists of a findings $x$, an indicator $i$, and a likely/less likely interpretation $y$ of the findings $x$ depending on $i$. 

\paragraph{Dataset Construction} We first find phrases such as ``likely represents'', ``consistent with'', and ``may be unrelated to'' that represent uncertainty from each sentence of reports. We view these phrases as indicators of the presence of interpretations; denote them by $s^+$ or $s^\sim$. A likely or less likely indicator (Appendix~\ref{sec:unification}) suggests a likely or less likely interpretation of a finding. For each likely indicator $s^+$, we treat the sub-sentence preceding $s^+$ concatenated with prior 6 sentences as the findings ${x}$, and the completion of the sentence following $s^+$ as the likely interpretation $y^+$ of the findings ${x}$. We include prior sentences to provide more context for reaching interpretations. For less likely indicators $s^\sim$, we treat the sub-sentence either following or preceding $s^\sim$ as the less likely interpretation of the findings depending on how $s^\sim$ is stated. An example can be found in Figure~\ref{fig:motivation_examples}.

\paragraph{Indicator Unification} We have collected a variety of indicators and decided to unify them into a minimum set for both likely and less likely indicators. More details of indicator unification can be found in Appendix~\ref{sec:unification}. 

\paragraph{Evaluation} \liyan{To ensure the human evaluation for \textsc{MRIInterpret} to be as reliable as possible, we carefully curate a thorough annotation instruction guideline with precise definitions for all relevancy labels in Section~\ref{sec:mri_human_eval} and Appendix~\ref{sec:annotation_brain}.}


\section{Evaluation on Commonsense Reasoning} \label{sec:result}

\subsection{Automatic Evaluation}

\begin{table}
\small
\centering
\renewcommand{\tabcolsep}{1.6mm} 
\begin{tabular}{lcccc}
\toprule
 & \multicolumn{2}{c}{ART} & \multicolumn{2}{c}{E-CARE} \\ 
  \cmidrule(r){2-3} \cmidrule(l){4-5}
\textbf{Model} & \textbf{Frac} ($\uparrow$) & \textbf{PPL} ($\downarrow$) & \textbf{Frac} ($\uparrow$) & \textbf{PPL} ($\downarrow$) \\
\midrule
\textsc{Mle} & 54.1 & 42.6 & 54.5 & 80.4 \\
\textsc{Mle-LL} & 56.6 & 42.5 & 52.6 & 84.8 \\
+ \textsc{CD} & 59.9 & 49.8 & \textbf{63.4} & 107.3 \\
+ \textsc{DExperts} & 56.2 & 51.7 & 57.2 & 108.3 \\
\midrule
\textsc{Brainstorm} & \textbf{79.4} & 40.7 & 58.1 & \textbf{69.2} \\
+ \textsc{CD} & \textbf{79.7} & 50.2 & \textbf{67.2} & 88.1 \\
 + \textsc{DExperts} & \textbf{79.0} & 51.5 & 58.1 & 89.3 \\
\midrule
\textsc{Quark} & \textbf{85.9} & \textbf{27.5} & \textbf{68.2} & 80.8\\
\midrule
\textsc{Brainstorm} &  &  &  &  \\
$-\mathcal{L_\mathrm{margin}}$ & \textbf{69.3} & 44.9 & 54.6 & 73.2 \\
$-\mathcal{L_\mathrm{sim}}$ & 58.2 & 52.6 & 53.2 & 83.7 \\
\textsc{Brainstorm}$'$ & 58.3 & 52.0 & 55.1 & 71.2 \\
\bottomrule
\end{tabular}
\caption{Performance of generating less likely hypothesis on \textsc{Art} test set and \textsc{E-CARE} validation set. For \textsc{DExperts} and \textsc{CD}, we list the fractions where models reach minimum PPL. The ablation study of our proposed method is shown at the bottom.}
\label{tab:compare_baseline}
\end{table}

Our first evaluation relies on automatically assessing whether system outputs are likely or less likely according to humans. We fine-tune DeBERTa models \cite{He2020DeBERTaDB} for our automatic evaluation on two everyday commonsense datasets. They take the pair of $(x, y)$ as input and predict whether $y$ is a likely or less likely hypothesis.
In our settings, the fine-tuned DeBERTa model achieves 85\% accuracy on the test set of \textsc{Art} and achieves 80\% on the original validation set of \textsc{E-CARE}.

Table~\ref{tab:compare_baseline} compares a number of methods on our commonsense reasoning datasets. 
We answer several questions based on these results. We perform a paired bootstrap test for each result by comparing to \textsc{Mle-LL}. We highlight results that are better at 0.05 level of significance.

\paragraph{Can we just train on $(x, y^\sim)$?} Interestingly, the baseline model \textsc{Mle-LL} that only trained on $(x, y^\sim$)  pairs generates ``likely'' hypotheses approximately half of the time. This is possibly an effect of the pre-training regimen; furthermore, generating likely hypotheses may be easier and past work has shown that seq2seq models can amplify behaviors like copying that are easy to learn \cite{goyal-etal-2022-training}.

\begin{figure*}
\small
    \centering
    \includegraphics[width=0.8\linewidth, trim=0mm 0mm 00mm 0mm,clip]{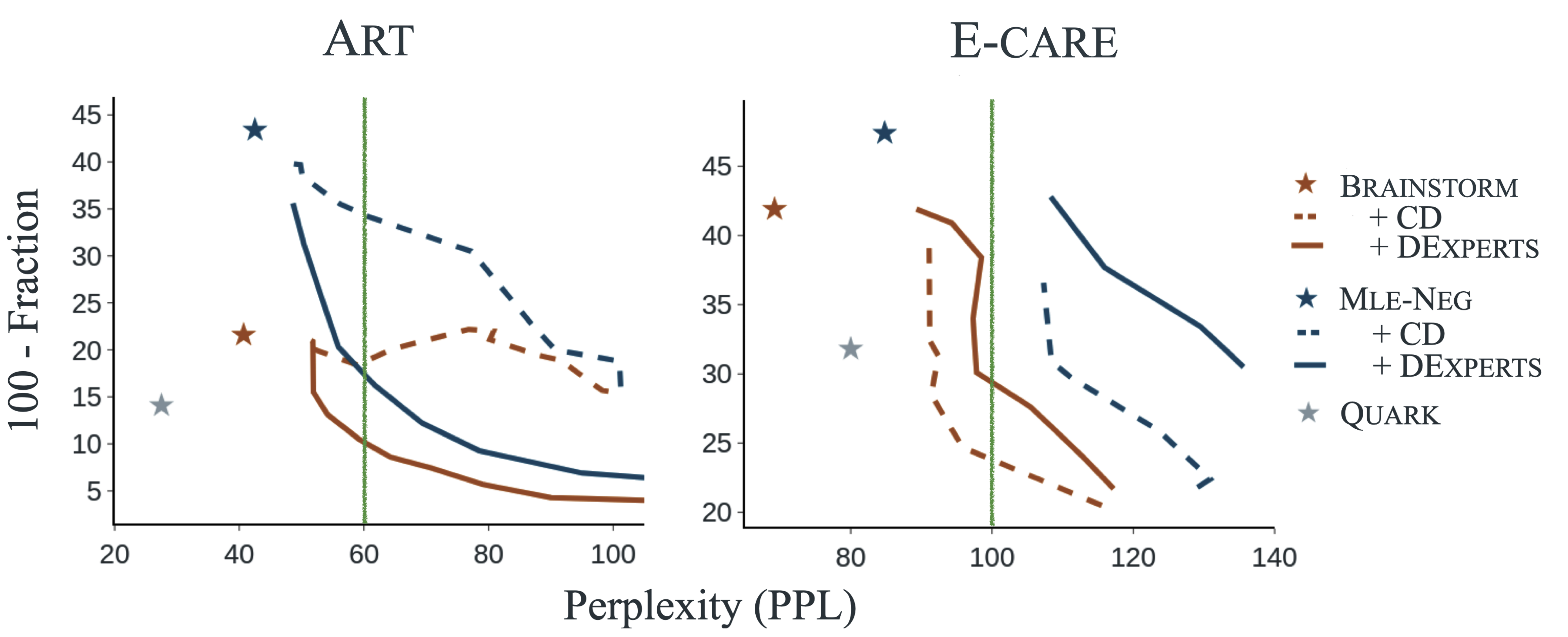}
    \caption{Fraction-perplexity trade-off of decoding-time methods \textsc{CD} and \textsc{DExperts} on \textsc{Art} test set and original \textsc{E-CARE} validation set (our test set). We show the trade-off across various values for $\tau_\mathrm{CD}$ in \textsc{CD} and $\alpha$ in DExperts. Both CD and DExperts can improve the fraction of less likely hypotheses, but at a very high cost to perplexity.}
    \label{fig:anlg_compare}
\end{figure*}



\begin{table*}[]
\centering
\small
\renewcommand{\tabcolsep}{1.8mm}
\begin{tabular}{lccccrccccrc}
\toprule
\multirow{3}{*}{\textbf{Model}} & \multicolumn{5}{c}{\textsc{Art}} & \multicolumn{5}{c}{\textsc{E-CARE}} \\
\cmidrule(r){2-6} \cmidrule(l){8-12}
 & \begin{tabular}[c]{@{}c@{}}\textbf{Likely}\\ \textbf{ ($\downarrow$)}\end{tabular} & \begin{tabular}[c]{@{}c@{}}\textbf{L-Likely} \\ \textbf{ ($\uparrow$)}\end{tabular} & \begin{tabular}[c]{@{}c@{}}\textbf{Contra.}\\ (?)\end{tabular} & \begin{tabular}[c]{@{}c@{}}\textbf{Rep.}\\ \textbf{($\downarrow$)}\end{tabular} & \begin{tabular}[c]{@{}c@{}}\textbf{Irrel.}\\ \textbf{($\downarrow$)}\end{tabular} && \begin{tabular}[c]{@{}c@{}}\textbf{Likely}\\ \textbf{ ($\downarrow$)}\end{tabular} & \begin{tabular}[c]{@{}c@{}}\textbf{L-Likely}\\ \textbf{ ($\uparrow$)}\end{tabular} & \begin{tabular}[c]{@{}c@{}}\textbf{Contra.}\\ (?)\end{tabular} &  \begin{tabular}[c]{@{}c@{}}\textbf{Rep.}\\ \textbf{($\downarrow$)}\end{tabular} & \begin{tabular}[c]{@{}c@{}}\textbf{Irrel.}\\ \textbf{($\downarrow$)}\end{tabular}\\
\midrule
\textsc{Mle-LL} & 42.3 & 15.2 & 22.7 & 9.5 & 10.3 && 35.4 & 15.6 & 5.7 & 18.6 & 24.7 \\
Quark & \textbf{14.7} & \textbf{20.8} & 51.0 & \textbf{4.3} & 9.2 && 35.2 & 15.1 & 5.7 & \textbf{3.3} & 40.7 \\
\textsc{Brainstorm} & \textbf{20.9} & \textbf{20.2} & 41.3 & \textbf{4.8} & 12.8 && 37.1 & \textbf{20.1} & 4.7 & \textbf{12.7} & 25.4 \\
\bottomrule
\end{tabular}
\caption{Human evaluations on \textsc{Art} and \textsc{E-CARE}. We see that our method is able to produce more ``less likely'' (L-Likely) outputs on both datasets. We calculated the mean of the ratings from multiple annotators for each sample.}
\label{tab:human_commonsense}
\end{table*}

\paragraph{Are the proposed two loss objectives effective?} We see that compared to \textsc{Mle-LL}, our proposed \textsc{Brainstorm} method achieves substantially higher fractions of less likely hypotheses with no cost to quality in terms of perplexity. At the bottom of Table~\ref{tab:compare_baseline}, we show that ablating either of the proposed loss objectives worsens performance (and note that ablating both yields \textsc{Mle}). \textsc{Brainstorm}$'$ is not as effective since it does not compare with outputs in the batch, but we can see its merits in \textsc{MRIInterpret} (Section~\ref{sec:mri_human_eval}).


\paragraph{Can decoding-time methods alleviate the problem of generating likely outputs?} We explore whether \textsc{DExperts} and \textsc{CD} can further raise the fraction of less likely generations when combined with either \textsc{Mle-LL} or \textsc{Brainstorm}. 
These methods have hyperparameters that trade off how much of the ``undesired'' behavior each can remove from the system. We compute several fraction-perplexity trade-off curves in Figure~\ref{fig:anlg_compare}. Notably, although the fraction of less likely outputs can improve, \textbf{both of these methods significantly increase the perplexity of generations}, which corresponds with notably worse fluency of the text. Although these points apparently have high less likely fractions, we caution that the distribution of the text may deviate from the text that DeBERTa was fine-tuned on, meaning that our classifiers may not work well in these ranges. The green lines reflect thresholds where we observe serious degradation in output quality starting to occur. 
Below this perplexity threshold, the automatic evaluation suggests that both methods demonstrate some capability in alleviating the models' tendency in generating ``likely'' hypotheses without too great a cost to perplexity. Note that \textsc{DExperts} is more effective than \textsc{CD} in \textsc{Art} and vice versa in \textsc{E-CARE}.

Table~\ref{tab:compare_baseline} reports the settings where models achieve the minimum perplexities; at these points, perplexity is substantially increased but the fraction of less likely hypotheses is not substantially changed for the majority of results.

\paragraph{Can \textsc{Quark} yield improvement?} In Table~\ref{tab:compare_baseline}, the automatic evaluation results show that \textsc{Quark} exceeds \textsc{Brainstorm} by generating 6\% more ``less likely'' hypothesis in \textsc{Art} and 10\% more in \textsc{E-CARE}. It also has lower perplexity in \textsc{Art}. To further compare the two models, we conducted a human evaluation on the outputs from two models, and the result shows that \textsc{Quark} generates lower-quality ``less likely'' hypotheses (Section~\ref{sec:human_eval_commonsense}).

\subsection{Human Evaluation} \label{sec:human_eval_commonsense}

To further validate the results, we conduct a finer-grained human evaluation on a sample of 100 examples from the test sets of both datasets along two axes -- relevancy and fluency. We refined our relevancy evaluation by dividing the ``relevancy'' category into four subcategories, resulting in a total of five categories for evaluation.: (1) \emph{Likely}; (2) \emph{Less likely}; (3) \emph{Contradictory} - the output is impossible if we assume the input is true; (4) \emph{Repetition} - the output is describing the same meaning as the input; and (5) \emph{Irrelevant} - the output has little connection with input. More thorough category definitions with examples, annotation instruction and quality checks for AMT annotators can be found in Appendix~\ref{sec:quality_check}. We compare the performance of three models: \textsc{Mle-LL}, \textsc{Brainstorm}, and \textsc{Quark} (Table~\ref{tab:human_commonsense}). As \textsc{Quark} demonstrates better performance in automatic evaluation, we include its generated text in our human evaluation.

Our results show a high level of agreement between the automatic evaluation (Table~\ref{tab:compare_baseline}) and human evaluation (Table~\ref{tab:human_commonsense}) regarding the fraction of ``likely'' hypotheses on both datasets. On \textsc{Art}, \textsc{Quark} and \textsc{Brainstorm} decrease the fraction of ``likely'' hypotheses by 60\% and 50\%, respectively, compared to \textsc{Mle-LL}. However, on \textsc{E-CARE}, the human evaluation indicates that all three models generate an equivalent number of ``likely'' hypotheses. 
By further breaking down the ``relevancy'' category used in the automatic evaluation, we then have a clearer understanding of the distribution of categories among the models' outputs.

\paragraph{Low-Quality Hypotheses} It is not desirable for models to generate outputs that are repetitions of the input (Repetition) or have little connection to the input (Irrelevant). On the \textsc{Art} dataset, all models generate a small proportion of irrelevant outputs, with \textsc{Quark} and \textsc{Brainstorm} reducing the fraction of ``Repetition'' hypotheses by half, compared to \textsc{Mle-LL}. However, we get more low-quality outputs on \textsc{E-CARE}. While \textsc{Brainstorm} is able to reduce the fraction of Repetition hypotheses by a large margin, it is not as effective as \textsc{Quark}. One possible reason for this is that \textsc{Quark} is trained to generate outputs that the DeBERTa classifier (the reward model) predicts as less likely; Repetition cases are rarely classified as less likely due to their similarity with the input, but Irrelevant outputs are more likely to be classified this way. 

\paragraph{Less Likely versus Contradictory} While less likely hypotheses are desirable, contradictory hypotheses are less so.
A typical way of generating a contradictory hypothesis is by simply adding negation: \emph{Lisa went laptop shopping yesterday} $\rightarrow$ \emph{Lisa \textbf{didn't} go laptop shopping yesterday}. However, such examples have little value as the negation brings no new information to the input and is not a useful counterfactual for a user to see. 

We evaluate the models' outputs on the \textsc{Art} dataset, where a significant number of contradictory hypotheses are generated, and find that 43 out of 100 hypotheses generated by \textsc{Quark} include the words ``didn't'' or ``not,'' while only 10 hypotheses generated by \textsc{Brainstorm} and \textsc{Mle-LL} did so. We posit that this is likely due to the DeBERTa classifier assigning high rewards for hypotheses that include negation words, and \textsc{Quark} effectively learning this shortcut.

\section{Human Evaluation on \textsc{MRIInterpret}} \label{sec:mri_human_eval}

To evaluate the models' performance on the radiological interpretation generation setting, we select 30 findings from our validation set that ask for less likely interpretation. For each finding, we select the human reference and generate the top 5 less likely interpretations from 2 baselines (\textsc{Mle-LL} and \textsc{Mle}) and \textsc{Brainstorm}$'$, resulting in 30 $\times$ (5 $\times$ 3 + 1) = 480 interpretations. We randomized the order of these interpretations before evaluation.

Due to the structure of the indicators in this dataset, methods that require examples to have both $y^+$ and $y^\sim$ for the same data (see ``pair'' in Table~\ref{tab:model_summary}) are not able to be used. Since \textsc{Quark} relies on a trained classifier, we choose not to use \textsc{Quark} as well. \liyan{A trained classifier on \textsc{MRIInterpret} is not reliable since the training set only consists of naturally occurring data, which is highly imbalanced (see Table~\ref{tab:stats} in Appendix). This leads the classifier to perform poorly on the ``less likely'' class, which is the minority class but is also the class of greatest interest in this study. We find that augmenting the training data with counterfactual cases is not easy. For example, ``\emph{the lack of evidence of restricted diffusion makes it less likely to be}'' is a naturally occurring prompt from a less likely example, and attempting to change it to a sentence such as ``\emph{the lack of evidence of restricted diffusion could represent}'' yields a statement that turns out to be out of distribution from the training data and models do not behave reliably in these counterfactual cases.}

\begin{table}[]
\small
\centering
\renewcommand{\tabcolsep}{2.0mm}
\begin{tabular}{@{}lcrrrr}
\toprule
\multirow{2}{*}{\textbf{Model}} & \multirow{2}{*}{ \textbf{Likely}} & \multicolumn{3}{c}{\textbf{Less likely}} & \multirow{2}{*}{\textbf{Irrel.}} \\
\cmidrule(r){3-5}
 &  & \textbf{High} & \textbf{Med.} & \textbf{Low} &  \\
\midrule
\textsc{Mle-LL} & 6.7 & 40.7 & 21.2 & 14.7 & 16.7 \\
\textsc{Mle} & 7.3 & 50.0 & 22.1 & 13.3 & 7.3 \\
\textsc{Brainstorm}$'$ & 6.7 & 42.0 & \textbf{32.6} & 8.7 & 10.0 \\
\midrule
Reference & 3.3 & 76.7 & 13.4 & 3.3 & 3.3 \\
\bottomrule
\end{tabular}
\caption{Human Evaluation on \textsc{MRIInterpret}. Results are shown as percentages. We evaluated 30 $\times$ 5 = 150 less likely interpretations generated from each model and 30 less likely interpretations from human reference. Results show that our proposed model successfully shifts the distribution of generated interpretations further toward the tail of the ``relevant but less likely'' category but still generates relevant diagnoses.}
\label{tab:human_eval_brain}
\end{table}



For each generated interpretation, we evaluate its (1) \textbf{relevancy} to the findings and (2) whether it contains any \textbf{hallucinations} about findings (Appendix~\ref{sec:brain_hallucination}). For relevancy, we asked a neurologist to classify each interpretation into: (1) \emph{Relevant and likely}; (2) \emph{Relevant and less likely}; and (3) \emph{Irrelevant}. Further, for those classified as ``Relevant and less likely'', we further evaluate how well the interpretation fits into the context of the findings by grading them on three levels: high, medium and low, ranging from high matches that represent the most obvious less likely interpretations to low matches that represent relevant but exceedingly rare diagnosis. We provide detailed definitions for these categories and include comprehensive annotation guidelines in Appendix~\ref{sec:annotation_brain} to facilitate consistency in future studies.

Results are shown in Table~\ref{tab:human_eval_brain}. Most human references (which the neurologist was blinded to) are annotated as either a high or medium match under the relevant but less likely category, suggesting the reliability of the neurologist's annotation. We find that training on all data (\textsc{Mle}) instead of exclusively on less likely data (\textsc{Mle-LL}) would effectively help generate more relevant but less likely interpretations and reduce the amount of irrelevant ones. One possible reason is that \textsc{MRIInterpret} is a highly imbalanced dataset (Table~\ref{tab:stats}).

By comparing the outcomes between human reference and \textsc{Brainstorm}, we find that \textsc{Brainstorm} tends to shift the distribution of generated interpretations towards generating lower matched interpretations, which effectively extends the beam of potential diagnoses that meet the criteria of ``relevant but less likely'' based on refuting findings. Anecdotally, interpretations in this medium category reflect the sort of alternative hypotheses and ``outside-the-box'' suggestions that represent the original goal of our approach.

\section{Conclusion}

In this work, we propose a new text generation task ``less likely brainstorming'' for reducing cognitive errors in interpreting findings of MRI reports. We found that simply training on less likely data does not help with generating less likely interpretations and hence propose a novel CL method to tackle the problem. In two settings, we show that our proposed training technique can effectively generate more ``less likely'' hypotheses, producing interpretations that radiologists may not think of, outperforming past training- and decode-time modifications to generation models.

\section*{Limitations}

Our brain MRI interpretations were evaluated by a single neurologist. Such annotations require deep expertise and are not easily carried out with high quality by trainees, which limited the amount of data we were able to collect. To ensure that the annotation would be as reliable as possible, we carefully thought of the dimensions in evaluating the generated interpretations and proposed a thorough annotation instruction guideline. We believe that future work can conduct more extensive studies using our annotation guidelines as a starting point. Further, the radiology reports we experiment with are from a single academic medical center, which makes the generalizability unclear. Future work is needed to evaluate the performance of our models on data from different medical centers. Finally, future work is needed to evaluate relevant and likely outputs from MRI interpretations to address different forms of interpretation bias and to expand the beam of potential likely diagnoses based on the findings.

Beyond the brain MRI interpretation experiments, our generation experiments are limited to a set of pre-trained models optimized for carrying out generation tasks in English. It is possible that multilingual models generating in languages other than English will show different properties. We are limited by the availability of resources for automatic evaluation in these settings, but a more extensive multilingual evaluation with human users could be conducted in the future.

\section*{Ethical Risks}

We are proposing better ways for incorporating systems into the radiological diagnostic process. This is aimed at helping improve human decision-making and mitigating the limitations of traditional fully-automatic approaches. However, we believe that it is imperative to rigorously test and evaluate these methods before they can be put into practical clinical settings. We are not claiming that these methods are ready for real-world adoption at this stage.

\section*{Acknowledgments}

We would like to thank Darcey Riley and TAUR lab at UT for discussion about DExperts and for providing feedback on this work. We acknowledge the funding support from National Science Foundation AI Center Institute for Foundations of Machine Learning (IFML) at University of Texas at Austin (NSF 2019844), as well as NSF CAREER Award IIS-2145280 and IIS-2145640, National Library of Medicine under Award No. 4R00LM013001, and a gift from Salesforce, Inc.

\bibliography{anthology,custom}
\bibliographystyle{acl_natbib}

\appendix

\section{Dataset statistics}

Dataset statistics can be found in Table~\ref{tab:stats}.

\begin{table*}[]
\centering
\small
\begin{tabular}{ccccccccc}
\toprule
\multirow{2}{*}{\textbf{Dataset}} & \multicolumn{4}{c}{\textbf{Train}} & \multicolumn{2}{c}{\textbf{Val}} & \multicolumn{2}{c}{\textbf{Test}} \\
\cmidrule(r){2-5} \cmidrule(r){6-7} \cmidrule(r){8-9}
 & \multicolumn{2}{c}{\textbf{Likely}} & \multicolumn{2}{c}{\textbf{Less Likely}} & \multicolumn{2}{c}{\textbf{Less Likely}} & \multicolumn{2}{c}{\textbf{Less Likely}} \\
 \midrule
\textsc{MRIInterpret} & \multicolumn{2}{c}{10097} & \multicolumn{2}{c}{1005} & \multicolumn{2}{c}{121} & \multicolumn{2}{c}{---} \\
\textsc{Art} & \multicolumn{2}{c}{50509} & \multicolumn{2}{c}{50509} & \multicolumn{2}{c}{1781} & \multicolumn{2}{c}{3562} \\
\multirow{2}{*}{\textsc{E-CARE}} & cause & effect & cause & effect & cause & effect & cause & effect \\
\cmidrule(r){2-3} \cmidrule(r){4-5} \cmidrule(r){6-7} \cmidrule(r){8-9}
 & 6855 & 6580 & 6855 & 6580 & 762 & 731 & 1088 & 1044 \\
 \bottomrule
\end{tabular}
\caption{A summary of dataset statistics. All datasets are in English. For \textsc{Art} and \textsc{E-CARE}, we show the stats of our adapted versions. Since \textsc{E-CARE} has a hidden test set, we randomly split the original training set into a training and a validation set, and we use the original validation set as our test set. Note that each example in \textsc{E-CARE} asks for either the cause or the effect of the premise.}
\label{tab:stats}
\end{table*}

\section{Definition of Relevancy Categories on Everyday Commonsense} \label{sec:definitions}

\liyan{To encourage more consistent human evaluations, we formally define all relevancy categories as the following. These definitions are refined from rounds of pilot studies to reduce ambiguity for human annotations. Example outputs and explanations for each relevancy category can be found in the annotation interface (Figure~\ref{fig:anlg_ins_1} and ~\ref{fig:ecare_ins_1}).}

\subsection{\textsc{E-CARE}}

\paragraph{Relevant} \liyan{A hypothesis is relevant if it fits with the same scenario as the premise. It should not introduce new people, places, or things that are not at least plausibly in the same source scenario.}

\paragraph{Likely} \liyan{For the hypothesis to be likely, it must also be causally related to the premise – either the premise causes the hypothesis or the hypothesis causes the premise (you will see both versions of the task below). There should not be clearly more likely hypotheses than it. }

\paragraph{Relevant and Less likely} \liyan{The hypothesis is still the same scenario as the premise (relevant). However, it is less likely to be causally related to the premise. There could be other hypotheses that are superior to the given hypothesis.}

\paragraph{Irrelevant} \liyan{The generated hypothesis does not describe the same scenario as the premise or is not causally related to the premise.}

\paragraph{Contradictory} \liyan{The hypothesis contradicts the premise – it says something that is impossible if we assume the premise to be true (e.g., the premise states that something happened and the hypothesis states that that thing did not happen). }

\paragraph{Repetition} \liyan{The hypothesis is very similar to the premise – it either contains a text span that is a repetition of the premise, or it is expressing nearly the same meaning as the premise.}

\subsection{\textsc{Art}}

\paragraph{Relevant} \liyan{A hypothesis is relevant if it fits with the same scenario as the observation pair. It should not introduce new people, places, or things that are not at least plausibly in the same source scenario.}

\paragraph{Likely} \liyan{For the hypothesis to be likely, it must also be strongly related to $O_1$ and $O_2$ in a causal fashion – to the extent possible, the first observation $O_1$ should cause the hypothesis and the hypothesis causes the second observation $O_2$. There should not be clearly more likely hypotheses than it.}

\paragraph{Relevant and Less likely} \liyan{The hypothesis is still the same scenario as the observation pair (relevant). However, it is less likely to be causally related to the observation pair – maybe it could happen following $O_1$, but not necessarily. There could be other hypotheses that are superior to the given hypothesis.}

\paragraph{Irrelevant} \liyan{The hypothesis does not describe the same scenario as the observation pair: it either involves different people, places, or things, or the events it describes have very little connection to $O_1$ and $O_2$.}

\paragraph{Contradictory} \liyan{The hypothesis contradicts either observation $O_1$ or observation $O_2$ – it says something that is impossible if we assume $O_1$ and $O_2$ to be true (e.g., $O_2$ states that something happened and the hypothesis states that that thing did not happen). }

\paragraph{Repetition} \liyan{The hypothesis is very similar to either $O_1$ or $O_2$ – it either contains a text span that is a repetition of $O_1$ or $O_2$, or it is expressing nearly the same meaning as $O_1$ or $O_2$. }

\section{Annotation on Everyday Commonsense} \label{sec:quality_check}
The human evaluation by crowdworkers has been judged to be IRB exempt. We hired crowd annotators from US through Amazon Mechanical Turk. These annotators have lifetime approval rates over 99\% and more than 1000 approved HITs. We first conducted a quality check on \textsc{Art} and \textsc{E-CARE}. For each dataset, we randomly selected 100 examples from the test set and each example is evaluated by 7 annotators, resulting in 100 $\times$ 7 = 700 annotations for each dataset. We finally selected 7 qualified crowdworkers from each of the datasets. The procedure of filtering out non-qualified workers is shown below. For qualified crowdworkers, we randomly select another 100 examples from each dataset and conduct a final annotation round, resulting in 100 $\times$ 7 $\times$ 2 = 1400 annotations in total. We set maximum time on completing each HIT to 1 hour and each HIT takes approximately 1.5 minutes. We paid annotators \$0.3/HIT, which is equivalent to \$12/hr and is higher than the minimum USA wage.

\begin{figure*}
    \centering
    \includegraphics[width=0.9\linewidth, trim=0mm 410mm 00mm 0mm,clip]{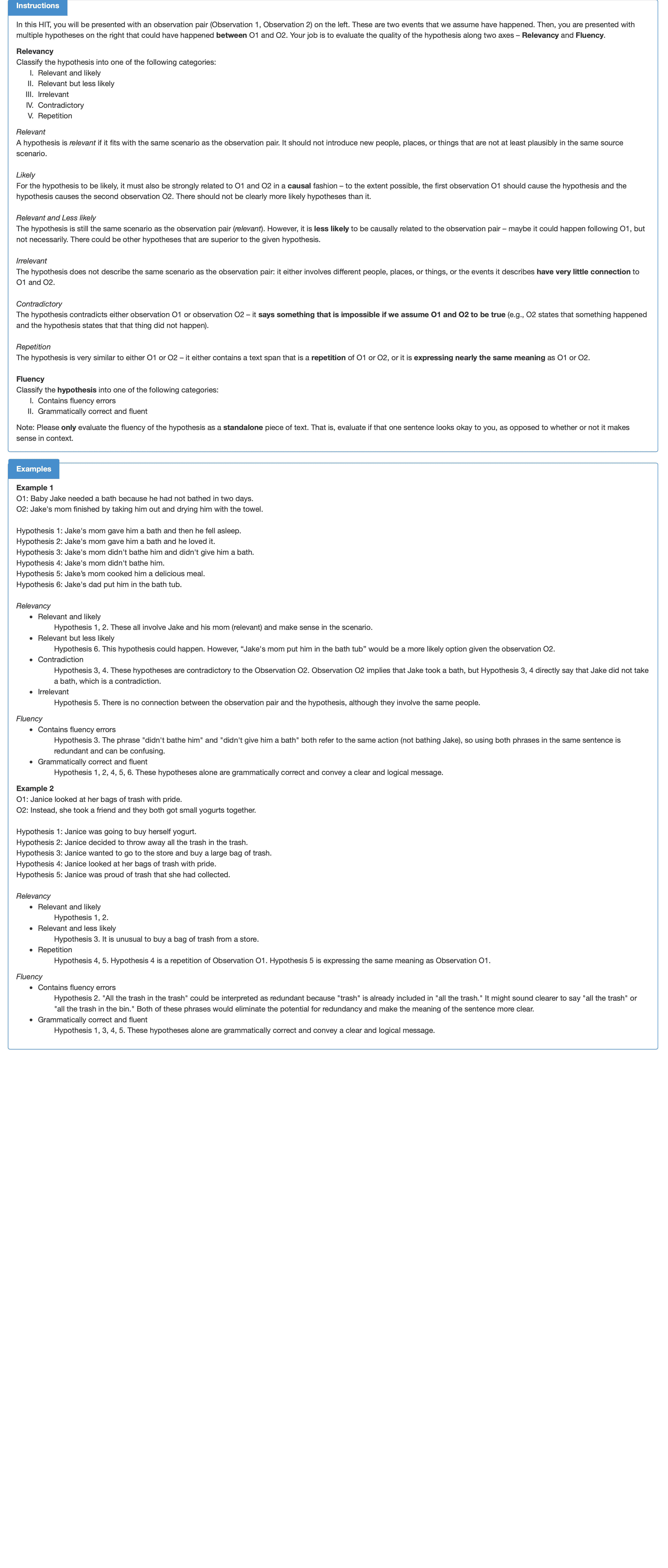}
    \caption{Annotation Interface (I) for \textsc{Art}.}
    \label{fig:anlg_ins_1}
\end{figure*}

\begin{figure*}
    \centering
    \includegraphics[width=0.9\linewidth, trim=0mm 0mm 20mm 0mm,clip]{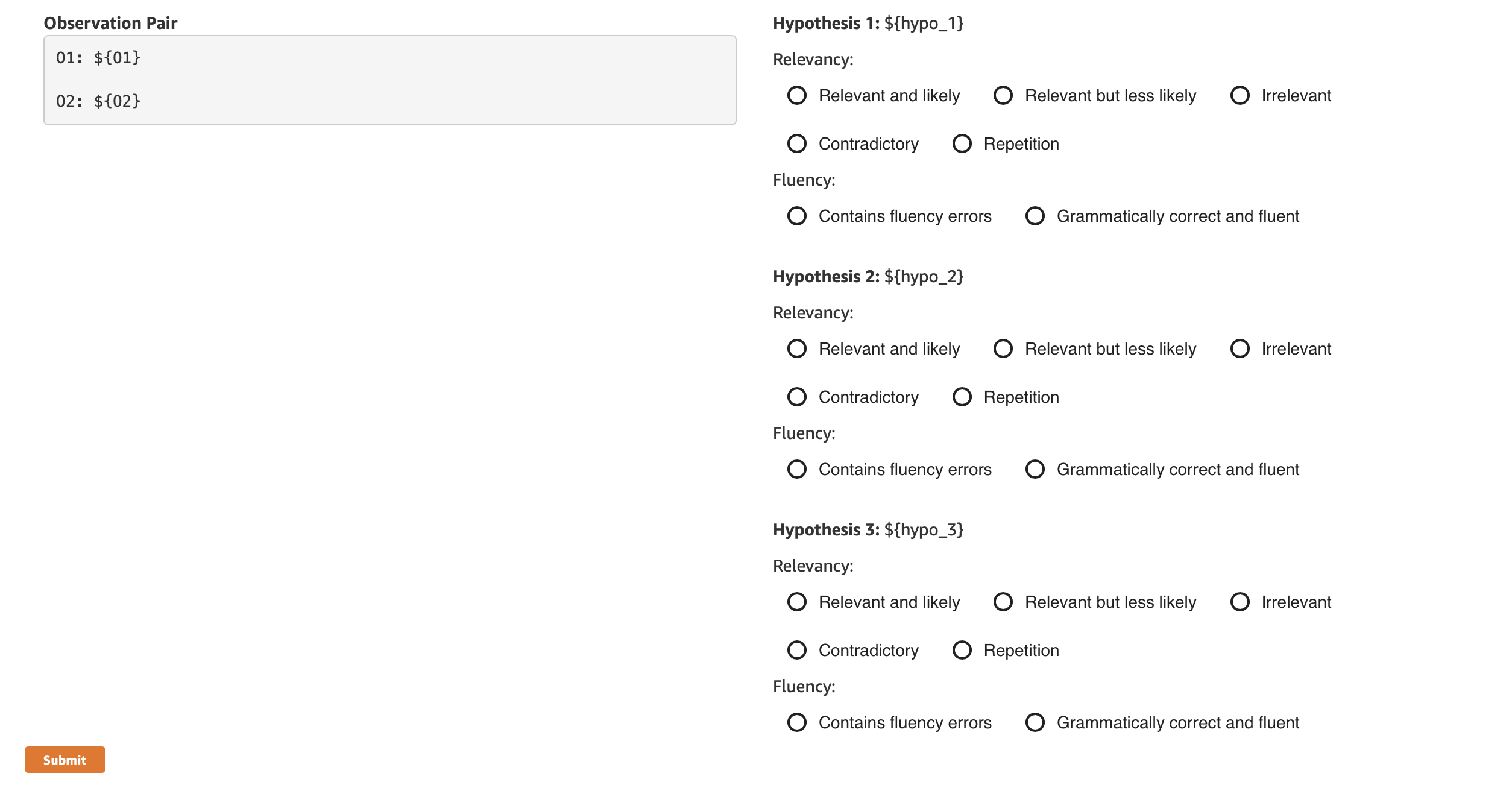}
    \caption{Annotation Interface (II) for \textsc{Art}.}
    \label{fig:anlg_ins_2}
\end{figure*}

\begin{figure*}
    \centering
    \includegraphics[width=0.9\linewidth, trim=0mm 410mm 00mm 0mm,clip]{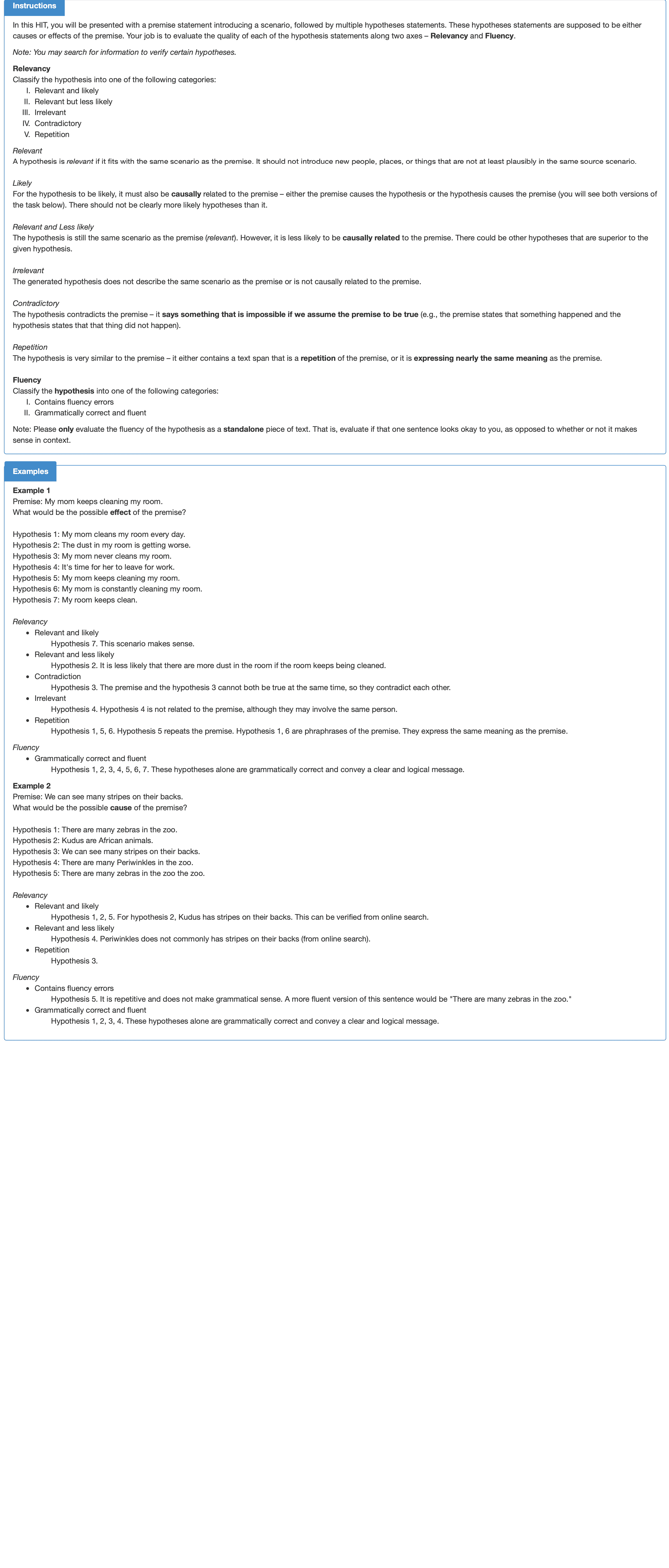}
    \caption{Annotation Interface (I) for \textsc{E-CARE}.}
    \label{fig:ecare_ins_1}
\end{figure*}

\begin{figure*}
    \centering
    \includegraphics[width=0.9\linewidth, trim=0mm 0mm 20mm 0mm,clip]{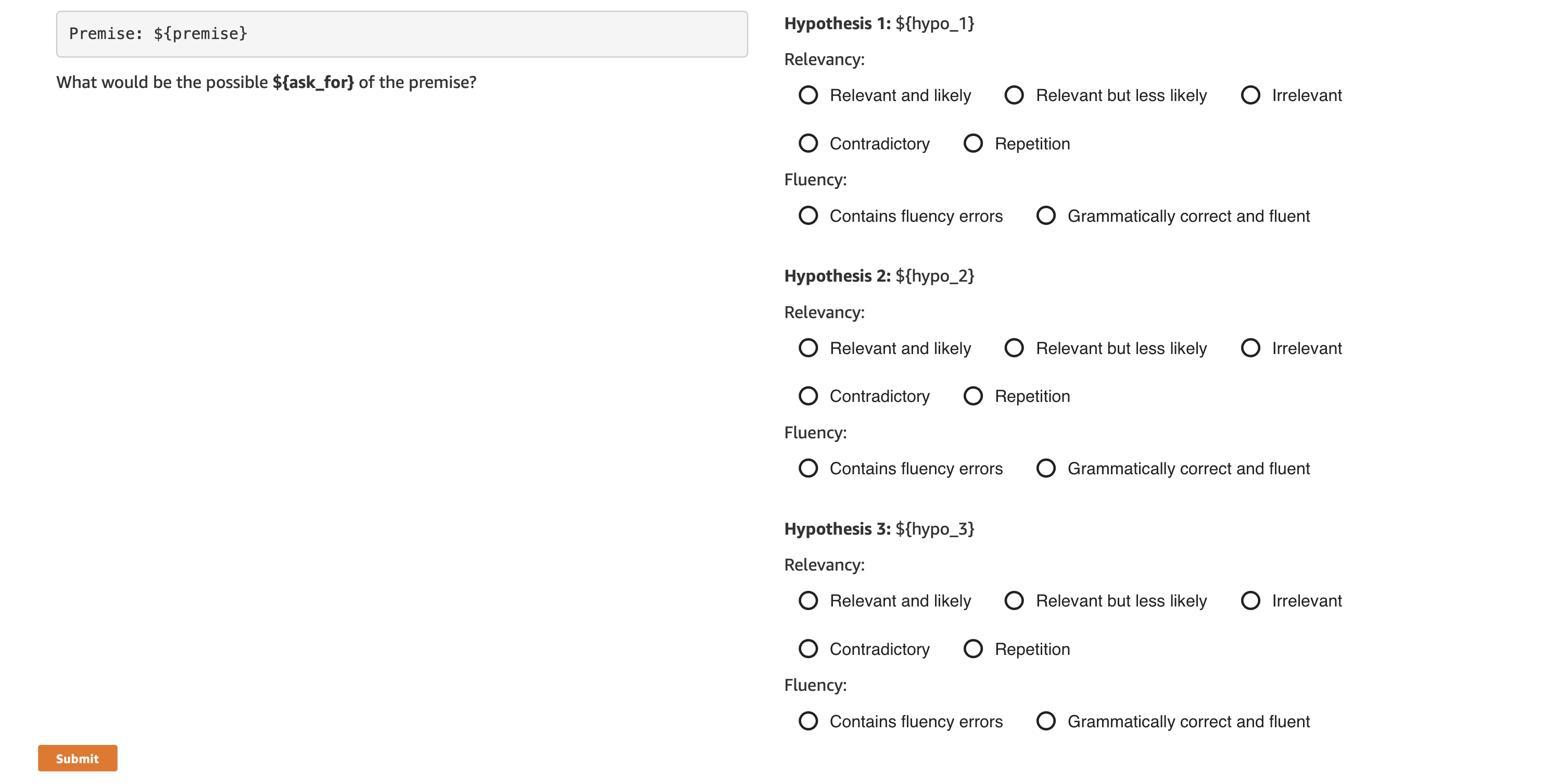}
    \caption{Annotation Interface (II) for \textsc{E-CARE}.}
    \label{fig:ecare_ins_2}
\end{figure*}

Category definitions and annotation instructions with examples are shown in Figure~\ref{fig:anlg_ins_1},~\ref{fig:anlg_ins_2},~\ref{fig:ecare_ins_1} and~\ref{fig:ecare_ins_2}.

\paragraph{Selecting Qualified Workers} After we collected all annotations from the pilot study. We filter out workers by following these steps:

\begin{enumerate}
    \item We first filter out workers that annotated less than 4 HITs. With limited amount of annotated HITs, it is hard to evaluate the consistency of their annotations.
    \item For any HIT, if two output sequences are exactly the same but the annotator assigned them different categories, then we remove the worker. For example, in \textsc{E-CARE}, if the premise is ``\emph{Tom goes to the gym every day.}'' and we have the hypotheses ``\emph{He gets a promotion from his manager who saw him in the gym.}'' that appears twice, then if one hypothesis is classified as ``Relevant and Likely'' and another one is classified as ``Relevant but Less Likely'', we will filter out this annotator.
    \item We use the ``Repetition'' category to further filter out annotators. We believe ``Repetition'' is the least subjective category in our annotation instruction, and using this category to filter out annotations would lead to minimum bias we can project to the selected annotators. This consists of two steps: (1) A model many generate an output that is exactly the input. For example, a model takes as input ``\emph{Tom goes to the gym every day.}'' and generate ``\emph{Tom goes to the gym every day.}'' as well. This happens sometimes across all models. For those cases, we will filter out annotators that assigned categories other than ``Repetition''; (2) Besides the exact match, there are cases where a model's output is a paraphrase of the input. For these, to minimize our bias, we choose to use models' outputs that only differs from the input by at most two words to filter out annotators. For example, in \textsc{Art}, if one observation is ``\emph{Lisa went laptop shopping yesterday}'', and the model's output is ``\emph{She went laptop shopping yesterday}'', then we filter out annotators that do not assign ``Repetition'' to it.
\end{enumerate} 

After we collected all the annotations from qualified workers, we use the above steps to further filter out works that do not meet our standard. Finally, we got valid annotations by three annotators from each datasets. We use Fleiss kappa to calculate the agreement between annotators. The annotators achieve moderate agreement ($\kappa$ = 0.447) on \textsc{Art} and fair agreement ($\kappa$ = 0.354) on \textsc{E-CARE} for relevancy evaluation. This is within our expectation since evaluating whether a hypothesis is likely or less likely is subjective.

\section{Fluency Evaluation on Everyday Commonsense Reasoning}

Fluency evaluation can be found in Table~\ref{tab:fluency_eval_commonsense}. Most of generations from models are fluent and grammatically correct.

\begin{table*}[]
\small
\centering
\begin{tabular}{ccc|cc}
 \toprule
\multirow{2}{*}{\textbf{Model}} & \multicolumn{2}{c}{\textsc{Art}} & \multicolumn{2}{c}{\textsc{E-CARE}} \\
\cmidrule(r){2-3} \cmidrule(r){4-5}
 & \begin{tabular}[c]{@{}c@{}}\textbf{Gram. Correct}\\ \textbf{Fluent}\end{tabular} & \begin{tabular}[c]{@{}c@{}}\textbf{Contain Flu}.\\ \textbf{Errors}\end{tabular} & \begin{tabular}[c]{@{}c@{}}\textbf{Gram. Correct}\\ \textbf{Fluent}\end{tabular} & \begin{tabular}[c]{@{}c@{}}\textbf{Contain Flu.}\\ \textbf{Errors}\end{tabular} \\
\midrule
\textsc{Mle-LL} & 93.9 & 6.1 & 99.0 & 1.0 \\
\textsc{Quark} & 94.6 & 5.4 & 98.0 & 2.0 \\
\textsc{Brainstorm} & 93.5 & 6.6 & 95.9 & 4.1 \\
\bottomrule
\end{tabular}
\caption{Human evaluation of fluency on everyday commonsense reasoning datasets. Annotators reached substantial agreement on both datasets. }
\label{tab:fluency_eval_commonsense}
\end{table*}

\begin{table}
\centering
\begin{tabular}{cc}
\toprule
\textbf{Model} & \textbf{Hallucination (\%)} \\
\midrule
\textsc{Mle-LL} & 23.3 \\
\textsc{Mle} & 30.0 \\
\textsc{Brainstorm} & 33.3 \\
\midrule
Reference & 6.6 \\
\bottomrule
\end{tabular}
\caption{Human evaluation on hallucinations. The result shows the percentage of hallucinations found in 150 generated interpretations from each model.}
\label{tab:brain_hallucination}
\end{table}

\begin{figure*}
    \centering
    \includegraphics[width=\linewidth, trim=0mm 0mm 00mm 0mm,clip]{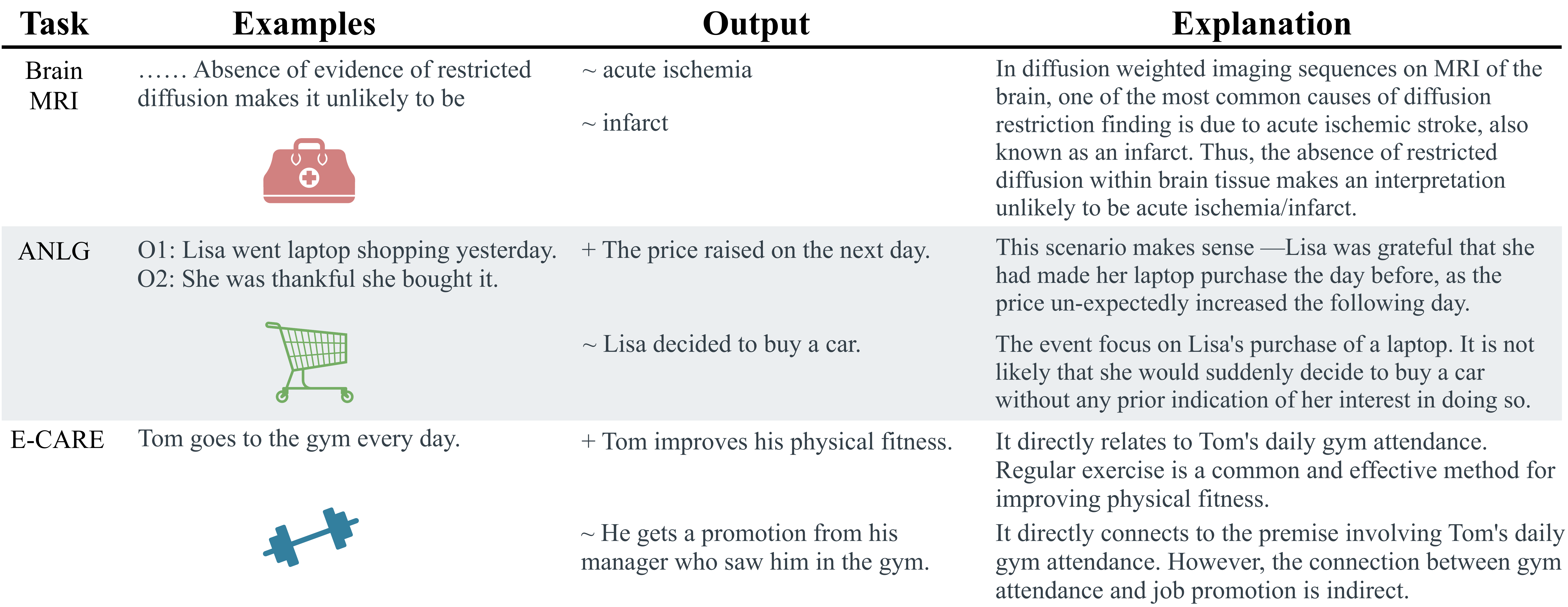}
    \caption{Examples from \textsc{MRIInterpret}, \textsc{Art} and \textsc{E-CARE}. The example shown in the table for \textsc{E-CARE} asks for a likely/less likely effect of the premise. ``$+$''/``$\sim$'' indicates whether humans would consider the output to be likely/less likely according to the context under the Examples column. We explain why humans would consider these outputs as likely/less likely in the Explanation column (this is not in the training data).}
    \label{fig:motivation_examples}
\end{figure*}

\begin{table*}[]
\centering\small
\begin{tabular}{ll}
\multicolumn{2}{l}{\begin{tabular}[c]{@{}l@{}}\textbf{O1:} Riley went to the store with her mother.\\ \textbf{O2:} Riley wore her cowboy boots to school the next day.\end{tabular}} \\
\toprule
\textsc{Mle-LL} & Riley’s mother bought her cowboy boots. \\
\quad + \textsc{CD} ($\tau_{\mathrm{CD}}$ = 0.5) & Riley had bought cowboy shoes that she had not worn before. \\
\quad + \textsc{CD} ($\tau_{\mathrm{CD}}$ = 1.0) & Her mother bought a new cowboy shirt for Riley. \\
\quad + \textsc{CD} ($\tau_{\mathrm{CD}}$ = 1.5) & Riiley got her new cowboy boots torn. \\
\quad + \textsc{DExperts} ($\alpha$ = 0.1) & Riley's mother bought her cowboy boots. \\
\quad + \textsc{DExperts} ($\alpha$ = 0.5) & Riley's mother bought her cowboy boots. \\
\quad + \textsc{DExperts} ($\alpha$ = 1.0) & Riley didn't have cowboy boots. \\
\\
\textsc{Brainstorm} & Riley bought a pair of cowboy boots for her mother. \\
\quad + \textsc{CD} ($\tau_{\mathrm{CD}}$ = 0.5) & Rice bought cowboy hats with her money and jeans. \\
\quad + \textsc{CD} ($\tau_{\mathrm{CD}}$ = 1.0) & Rice's mother bought her a pair of cowboy boots to wear to school next to her cowboy outfit. \\
\quad + \textsc{CD} ($\tau_{\mathrm{CD}}$ = 1.5) & Her mom bought her a new pair of cowboy boots. \\
\quad + \textsc{DExperts} ($\alpha$ = 0.1) & Riley's mother bought her a pair of cowboy boots. Riley hated them. \\
\quad + \textsc{DExperts} ($\alpha$ = 0.5) & Riley bought a pair of cowboy pants. \\
\quad+ \textsc{DExperts} ($\alpha$ = 1.0) & Riley bought a pair of cowboy pants. \\
\\
\textsc{Quark} & Riley's mother bought her cowboy boots at the store and Riley hated them. \\
\bottomrule
\end{tabular}
\caption{Examples of generating less likely hypotheses across different methods on \textsc{Art}.}
\label{tab:generation_anlg}
\end{table*}

\begin{table*}[]
\centering\small
\begin{tabular}{ll}
\multicolumn{2}{l}{\begin{tabular}[c]{@{}l@{}}\textbf{Premise:} The whole kitchen was burnt down more quickly.\\ \textbf{Ask for: less likely cause}\end{tabular}} \\
\toprule
\textsc{Mle-LL} & The whole kitchen was burnt down more quickly. \\
\quad + \textsc{CD} ($\tau_{\mathrm{CD}}$ = 0.5) & Tom poured water in the pan of acid. \\
\quad + \textsc{CD} ($\tau_{\mathrm{CD}}$ = 1.0) & There was a big thunderstorm yesterday. \\
\quad + \textsc{CD} ($\tau_{\mathrm{CD}}$ = 1.5) & There is a big storm outside. \\
\quad + \textsc{DExperts} ($\alpha$ = 0.1) & The whole kitchen was burnt down more quickly. \\
\quad + \textsc{DExperts} ($\alpha$ = 0.5) & The fire in the kitchen spread outwards. \\
\quad + \textsc{DExperts} ($\alpha$ = 1.0) & There are more and more fires in this place. \\
\\
\textsc{Brainstorm} & Tom put a lot of fuel on the fire. \\
\quad + \textsc{CD} ($\tau_{\mathrm{CD}}$ = 0.5) & Tom poured a bucket of water to a sink which has a high temperature. \\
\quad + \textsc{CD} ($\tau_{\mathrm{CD}}$ = 1.0) & There was an accident at night. \\
\quad + \textsc{CD} ($\tau_{\mathrm{CD}}$ = 1.5) & Tom poured gasoline to the stove. \\
\quad + \textsc{DExperts} ($\alpha$ = 0.1) & There is a fire in the kitchen. \\
\quad + \textsc{DExperts} ($\alpha$ = 0.5) & The whole kitchen was filled with smoke. \\
\quad+ \textsc{DExperts} ($\alpha$ = 1.0) & Tom's kitchen is leaking water. \\
\\
\textsc{Quark} & The fire in the kitchen was very hot. \\
\bottomrule
\end{tabular}
\caption{Examples of generating less likely hypotheses across different methods on \textsc{E-CARE}.}
\label{tab:generation_ecare}
\end{table*}

\section{Annotation on Brain MRI Interpretation} \label{sec:annotation_brain}

The use of the brain MRI data is covered by an IRB. A neurologist reviewed each finding sample and evaluated the interpretation on multiple metrics.

\subsection{Relevancy}

The overall objective of the interpretation generation was to produce less likely diagnoses, or interpretations, based on the absence of specific findings. The findings followed a common pattern of ``Absence of [finding x] makes it unlikely to be [interpretation y].'' The finding of interest was modified to be standardized across all findings if it used varying terminologies in a similar pattern (see Appendix~\ref{sec:unification} for more details). Because the interpretations are oriented in this negated valence, the objective of the output is to produce ``relevant but unlikely'' interpretations. The annotator rated the interpretation on 3 metrics: (1) relevant and likely, (2) relevant but less likely, and (3) irrelevant. 

\paragraph{Relevant and Likely} Output was judged as ``relevant and likely'' if the interpretation erroneously suggested a diagnosis that would be likely, not unlikely, despite the absence of [finding x]. For instance, ``Absence of restricted diffusion within the previously described fluid collections along the right convexity makes it unlikely to be''. An interpretation of ``the presence of a small subdural hematoma'' is actually a likely diagnosis given the lack of restricted diffusion in the fluid collection since subdural hematomas do not normally demonstrate restricted diffusion.

\paragraph{Relevant but Less Likely} Output was judged as ``relevant but less likely'' if the interpretation correctly provides a less likely diagnosis due to the absence of [finding x]. For example, ``absence of restricted diffusion makes it unlikely to be''. An interpretation of ``acute ischemia'' is unlikely since diffusion restriction is often associated with acute ischemia.

If the interpretation was judged as ``relevant but unlikely'', the degree to which the interpretation fits with the findings was graded on three levels: (1) high, (2) medium, and (3) low.

\begin{itemize}
    \item Less likely interpretations were \textbf{high matches} if they were within the top 5 diagnoses to fit the statement. These were the most obvious interpretations.
    \item Less likely interpretations were \textbf{medium matches} if they were further down the bar of potential interpretations. They still were relevant to the findings and made sense as being less likely given the absence of the finding of interest, but are less obvious and fall outside of the top 5 diagnoses.
    \item Less likely interpretations were \textbf{low matches} if the interpretation was relevant to the findings, but was an exceedingly rare diagnosis to make it of low value to mention as an interpretation.
\end{itemize}

\paragraph{Irrelevant} Output was judged as ``irrelevant'' if it was not related to the finding of interest or the structure that the finding of interest is referring to.

\subsection{Presence of Hallucination} \label{sec:brain_hallucination}

Lastly, no matter the rating of relevance, presence or absence of hallucination was noted. It was possible to have a relevant but unlikely interpretation with high degree of fit with the finding, but a hallucination that does not appear in the original findings was added. We therefore evaluate whether each interpretation contains hallucinations.

The results are shown in Table~\ref{tab:brain_hallucination}. The models listed contain a large proportion of hallucinated content especially for \textsc{Mle} and \textsc{Brainstorm}. We examined what these hallucinations look like. We found that in the most cases, models hallucinate about the findings (generating some findings that do not actually written in the report) and concatenate those hallucinated findings after their interpretations. For examples, a generated interpretation would be ``an acute infarction \emph{although this is limited by the presence of contrast enhancement}'',  ``intracranial abscess \emph{although this is limited by the presence of significant soft tissue swelling}'', or ``blood products in the ventricular system \emph{as seen on prior CT}.'' 

However, unlike other text generation tasks such as text summarization where hallucinations are hard to identify, hallucinations in \textsc{MRIInterpret} follow a pattern of interpretation followed by the non-existent findings. Although future work could work on how to directly generate interpretations without hallucination, a rule-based heuristics can remove the majority of hallucinations in the current version of our system. 

\section{Indicator Unification for \textsc{MRIInterpret}} \label{sec:unification}

\begin{figure*}
    \centering
    \includegraphics[width=\linewidth, trim=0mm 0mm 00mm 0mm,clip]{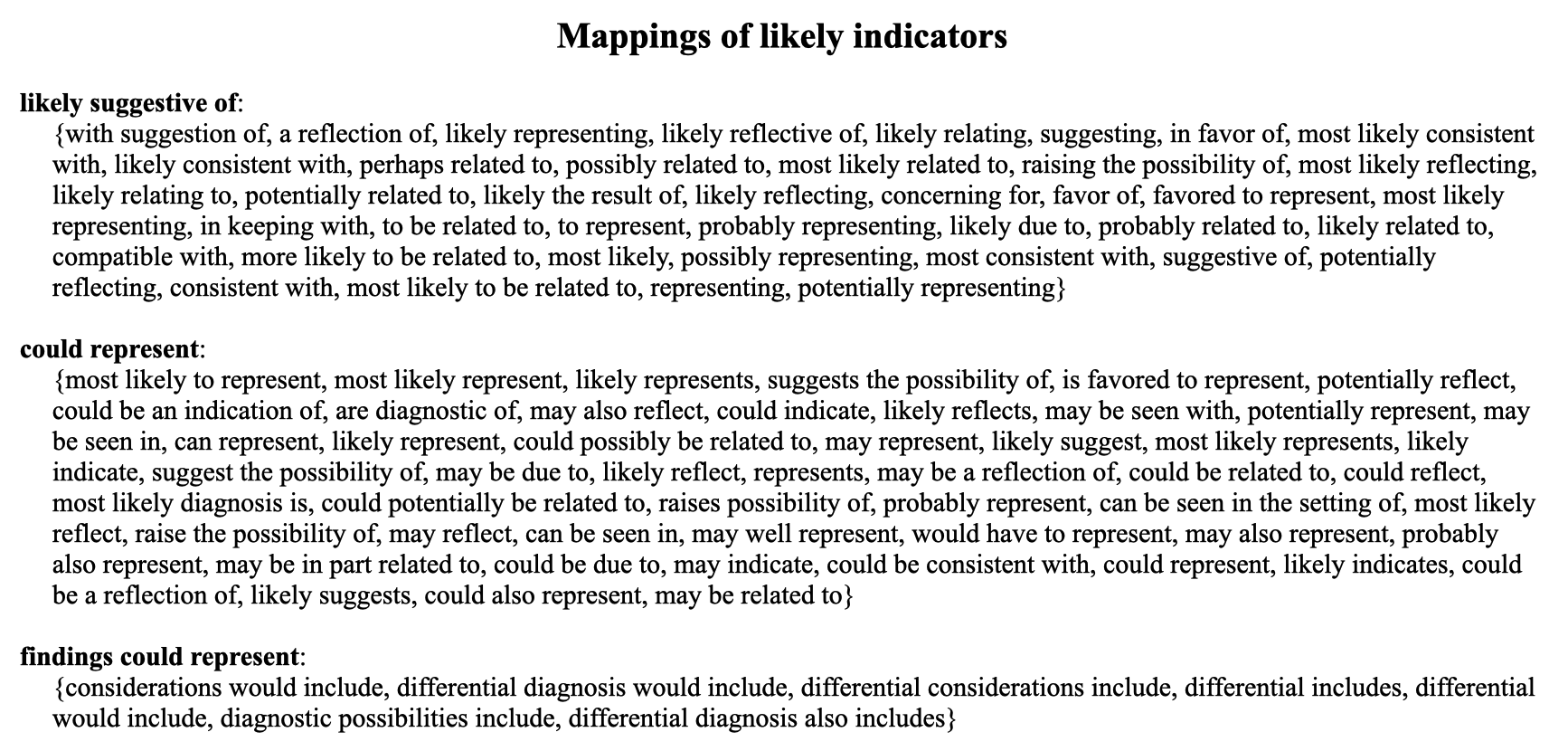}
    \caption{Unifying ``likely'' indicators in \textsc{MRIInterpret}.}
    \label{fig:unification_1}
\end{figure*}

\begin{figure*}
    \centering
    \includegraphics[width=\linewidth, trim=0mm 0mm 00mm 0mm,clip]{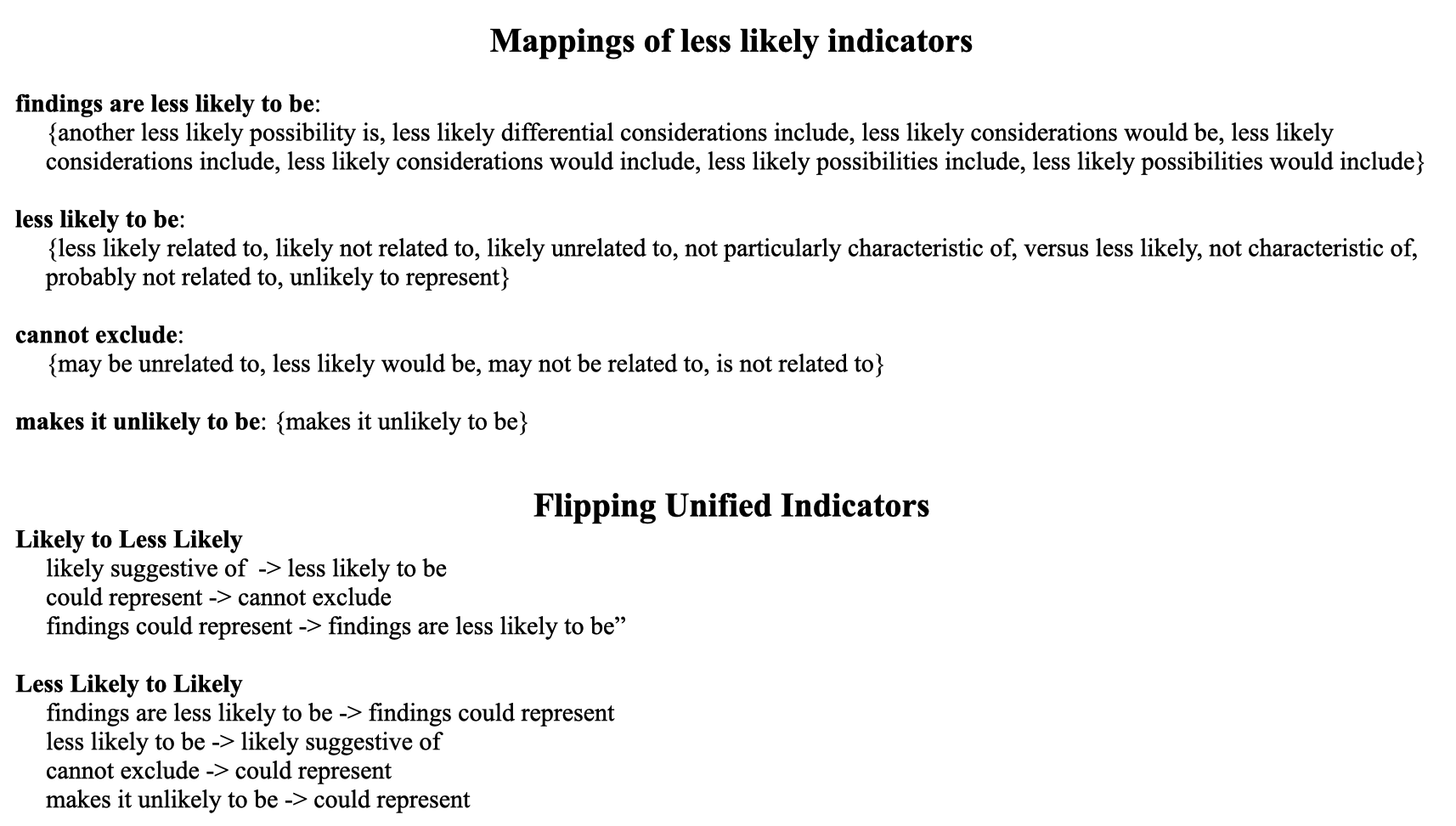}
    \caption{Unifying ``less likely'' indicators in \textsc{MRIInterpret} and how we map flipped indicators.}
    \label{fig:unification_2}
\end{figure*}

We narrowed down the indicators to a smaller set to ensure that our model sees sufficient data for each indicator during training. The indicator mappings are shown in Figure~\ref{fig:unification_1} and~\ref{fig:unification_2}. We also include the way we flip these indicators for the margin loss objective. 

\section{Example of generated outputs}

We show examples of generated outputs for both everyday commonsense reasoning datasets in Table~\ref{tab:generation_anlg} and~\ref{tab:generation_ecare}.

\section{Implementation Details}

\subsection{Significance Test}

 We perform a paired bootstrap test for each result by comparing to \textsc{Mle-LL}. We highlight results that are better at 0.05 level of significance.

\subsection{Computing Infrastructure} 

We use BART from HuggingFace Transformers \cite{wolf-etal-2020-transformers}, which is implemented in the PyTorch framework.

\subsection{Training Details} 

We fine-tune BART-Large (400M parameters) with 1 NVIDIA RTX A6000 GPU on all experiments and it converges in 2 epochs. We use AdamW as our optimizer with adam epsilon set to 1e-8. Learning rate is set to 5e-5 with linear schedule warmup. There is no warm-up step.

\subsubsection{Everyday Commomsense Reasoning}

We initialize the model from {\fontfamily{qcr}\selectfont facebook/bart-large}. The batch size is set to 64 if only using MLE objective and 42 otherwise. We set maximum input length to 100 and maximum output length to 64. Most text should fit into these lengths. The average training time for each model is around 0.8 GPU hours if only using MLE objective and 1.5 GPU hours otherwise.

\subsubsection{\textsc{MRIInterpret}}

We initialize the model from {\fontfamily{qcr}\selectfont GanjinZero/biobart-large} \cite{BioBART}. The batch size is set to 32. We set maximum input length to 256 and maximum output length to 60. Most text should fit into these lengths. The average training time for each model is around 0.8 GPU hours if only using MLE objective and 1.2 GPU hours otherwise.

\subsection{Hyperparameter Setups}

\paragraph{\textsc{Brainstorm}} For the margin loss $\mathcal{L_{\mathrm{margin}}}$ (Equation~(\ref{equ:margin_loss})), we chose $m$ within in the range of $1 \times 10 ^{-3}$ and $1 \times 10 ^{-2}$ and set it to 0.005 in the $\log$ space as it works well throughout our experiments. $w_s$ and $w_m$ are set to 1.0 and 10.0, respectively, as they achieve the best result on the validation set.  

\paragraph{\textsc{Quark}} We follows the default parameter setups in the original work with 6000 training steps for both commonsense reasoning datasets.

\paragraph{Decoding} We use diverse beam search for all experiments with diversity penalty set to 1.0. We set $\tau_\mathrm{CD}$ in \textsc{CD} from $2 \times 10 ^{-1}$ to $1 \times 10 ^{3}$, and $\alpha$ in \textsc{DExperts} from $1 \times 10 ^{-3}$ to $1$. We keep the recommended values for the remaining hyperparameters.

\end{document}